\documentclass[a4paper]{article}

\usepackage[letterpaper]{geometry}
\usepackage[parfill]{parskip}
\PassOptionsToPackage{numbers, compress}{natbib}
\usepackage[numbers, compress]{natbib}
\usepackage{xr-hyper,refcount}

\usepackage{graphicx}
\usepackage[utf8]{inputenc} 
\usepackage[T1]{fontenc}    
\usepackage{url}  
\usepackage{tabularx}
\usepackage{booktabs}       
\usepackage{amsfonts}       
\usepackage{nicefrac}       
\usepackage{microtype}      
\usepackage[parfill]{parskip}
\usepackage{algorithm,algorithmic}
\usepackage{amsmath,amsthm,amssymb,bbm}
\usepackage{mathtools}
\usepackage{cases}
\usepackage{comment}
\usepackage{subcaption}
\usepackage{algorithm,algorithmic}
\usepackage{color}
\usepackage{appendix}
\usepackage{xspace}
\usepackage{enumitem}
\usepackage[english]{babel}
\usepackage{authblk}

\usepackage[colorlinks,citecolor=blue,urlcolor=blue,linkcolor=blue,linktocpage=true]{hyperref}
\usepackage{cleveref}
\crefformat{equation}{(#2#1#3)}
\crefrangeformat{equation}{(#3#1#4) to~(#5#2#6)}
\crefname{equation}{}{}
\Crefname{equation}{}{}

\crefname{definition}{\textbf{definition}}{definitions}
\Crefname{definition}{Definition}{Definitions}
\crefname{assumption}{\textbf{assumption}}{assumptions}
\Crefname{assumption}{Assumption}{Assumptions}
\definecolor{maroon}{RGB}{192,80,77}

\usepackage[utf8]{inputenc} 
\usepackage[T1]{fontenc}    
\usepackage{hyperref}       
\usepackage{url}            
\usepackage{booktabs}       
\usepackage{amsfonts}       
\usepackage{nicefrac}       
\usepackage{microtype}      
\usepackage{tikz}
\usepackage{tabularx}
\usetikzlibrary{automata,decorations.markings,positioning,arrows}

\usepackage{natbib}
\usepackage{amsmath}
\usepackage{bbm}

\usepackage[normalem]{ulem}

\usepackage{mathtools}

\usepackage{amssymb,amsmath,amsthm}
\usepackage{soul}
\usepackage{url}            
\usepackage{xcolor}
\usepackage{xr-hyper,refcount}
\usepackage[colorlinks]{hyperref}  

\newcommand{\mhe}{$\lambda_{\max}$}

\title{On the Maximum Hessian Eigenvalue and Generalization}

\begin{document}

\author{
  Simran Kaur$^{\dagger}$, Jeremy Cohen$^{\dagger}$, Zachary C. Lipton$^\dagger$\\
  $^\dagger$Carnegie Mellon University\\
  \{\href{mailto:skaur@andrew.cmu.edu}{\nolinkurl{skaur}}, \href{mailto:jeremycohen@cmu.edu}{\nolinkurl{jeremycohen}},
  \href{mailto:zlipton@cmu.edu}{\nolinkurl{zlipton}}\}@cmu.edu
}

\maketitle

\begin{abstract}
The mechanisms by which
certain training interventions, 
such as increasing learning rates 
and applying batch normalization,
improve the generalization of deep networks
remains a mystery.
Prior works have speculated that
``flatter'' solutions generalize better 
than ``sharper'' solutions to unseen data,
motivating several metrics for measuring flatness
(particularly \mhe, the largest eigenvalue of the Hessian of the loss);
and algorithms, such as \textit{Sharpness-Aware Minimization} (SAM) 
\citep{foret2020sharpness}, that \emph{directly} optimize for flatness.
Other works question the link between \mhe\ 
and generalization.
In this paper, we present findings that call 
\mhe's influence on generalization
further into question.
We show that:
(1) while larger learning rates 
reduce \mhe\
\emph{for all batch sizes}, 
generalization benefits sometimes vanish at larger batch sizes;
(2) by scaling batch size and learning rate 
simultaneously,
we can change \mhe\ 
without affecting generalization;
(3) while SAM produces smaller \mhe\ \emph{for all batch sizes},
generalization benefits (also) vanish with larger batch sizes;
(4) for dropout, excessively high dropout probabilities can degrade generalization,
even as they promote smaller \mhe;
and (5) while batch-normalization does not
consistently produce smaller \mhe,
it nevertheless confers generalization benefits.
While our experiments affirm the generalization benefits 
of large learning rates and SAM for minibatch SGD,
the GD-SGD discrepancy demonstrates limits 
to \mhe's ability 
to explain generalization in neural networks.
\end{abstract}
\section{Introduction}
\label{sec:introduction}
Researchers have devised a repertoire of practices 
that reliably improve the generalization error of neural networks
\citep{DBLP:journals/corr/ZhangBHRV16,DBLP:journals/corr/Smith15a,smith2018dont,NEURIPS2019_dc6a7071,DBLP:journals/corr/abs-2005-07360,DBLP:journals/corr/IoffeS15,JMLR:v15:srivastava14a,jastrzebski2018factors,keskar2017largebatch}.
Techniques linked to improved test accuracy
across a wide variety of datasets and architectures 
include (i) small batch sizes for minibatch SGD; 
(ii) larger learning rates for minibatch SGD;
(iii) batch normalization;
and (iv) dropout; among others
\citep{NEURIPS2019_dc6a7071,DBLP:journals/corr/IoffeS15,JMLR:v15:srivastava14a,jastrzebski2018factors,keskar2017largebatch}.
Absent of guiding theory to explain \textit{why} 
these training interventions improve generalization,
researchers are left to rely on intuition.
One popular claim in the literature is that 
\emph{flat minima} generalize better 
than \emph{sharp} (non-flat) \emph{minima} 
\citep{hochreiter1996, keskar2017largebatch}.
Thus motivated, \citet{foret2020sharpness} proposed 
\emph{Sharpness-Aware Minimization} (SAM),
which directly optimizes for flat solutions. 

While there are many mathematical definitions of flatness, 
a common one is  
the largest eigenvalue of the Hessian of the loss with respect to the model parameters on the training set,
which we denote \mhe\ 
\citep{jastrzebski2018factors, lewkowycz2020large, dinh2017sharp}.
Already in the literature, there are indications that \mhe\ 
may not be causally linked with improved generalization.
In particular, \citet{dinh2017sharp} showed 
that given any set of weights for a ReLU network,
there exists a new set of weights 
that produce the same function, 
but have arbitrarily high \mhe.
However, a motivated reader might still argue 
that \citet{dinh2017sharp}'s construction is artificial: 
perhaps standard optimizers and initializations
do not give rise to pathologically sharp solutions.
Thus, while \citet{dinh2017sharp} convincingly demonstrates 
that ``high \mhe'' does not causally imply ``bad generalization,'' 
it leaves open the possibility that \mhe\ 
might still be a useful predictor of generalization 
in realistic deep learning scenarios.

In this paper, we conduct an experimental study to investigate
the association between \mhe\ and generalization.
Our results call the relationship
between \mhe\ and generalization 
further into question.
We provide four examples of training interventions 
that promote smaller values for \mhe,
but either do not change, or even degrade, generalization.
First, we show that for a non-batch-normalized VGG network \emph{in the large-batch setting}, 
increasing the learning rate reliably results in a smaller \mhe, 
yet does not improve generalization.
Second, we show that for the same architecture in the \emph{small-batch setting}, 
increasing the learning rate and batch size in tandem while preserving their ratio 
reliably results in a smaller \mhe, 
yet leaves generalization unchanged
(due to the linear scaling rule \citep{DBLP:journals/corr/GoyalDGNWKTJH17}).
Third, we show that for all batch sizes, 
a higher sharpness penalty in the context of SAM 
produces smaller \mhe.
However, SAM's generalization benefits 
diminish as batch size increases, 
vanishing entirely in the large-batch setting.
(In line with \citet{foret2020sharpness},
we observe that in the \textit{small-batch} setting, 
SAM promotes both flatness and generalization.) 
Fourth, we find that while dropout promotes smaller \mhe,
excessively high dropout probabilities degrade generalization.
In these four instances, 
promoting smaller \mhe\
does not reliably boost generalization.
Our results suggest that \mhe\ may not be reliably correlated with generalization in practice. 

We also provide an example of an intervention---batch-normalization---that 
improves generalization, without changing \mhe.
In particular, we find that while adding batch-normalization 
does not result in smaller \mhe\ in the large-batch setting, 
it does confer generalization benefits at large learning rates.
This finding further calls into question 
the correlation of \mhe\ with generalization. 

\section{Background and Related Work}
\label{sec:relatedwork}
In deep learning, certain design choices,
such as batch size and learning rate, 
are known to have an implicit
effect on generalization \citep{jastrzebski2018factors, NEURIPS2019_dc6a7071,keskar2017largebatch}.  
These choices are believed to implicitly guide the training process towards solutions with favorable geometric properties
\citep{jastrzebski2018factors, smith2018bayesian, lewkowycz2020large}.
These favorable geometric properties are broadly referred to as ``flatness'' 
(versus ``sharpness'').

\textbf{Metrics for flatness \enspace}
Some metrics, such as the maximum Hessian eigenvalue, 
measure the \emph{worst-case} loss 
increase under an \emph{adversarial} perturbation to the weights
 \citep{keskar2017largebatch, jiang2019fantastic},
while other proposed metrics,
such as the Hessian trace,
measure the \emph{expected} loss increase 
under \emph{random} perturbations to the weights. 
Finally, there are volume-based metrics, which can be defined as a large connected region in the weight space where the error remains approximately constant \citep{hochreiter1996}.
In this paper, we focus exclusively
on the leading Hessian eigenvalue \mhe\
and its 
relationship to
generalization.

\textbf{$\mathbf{\lambda_{max}}$ as a metric for flatness \enspace}
The maximum Hessian eigenvalue \mhe\ is commonly viewed as a potential sharpness/flatness metric
\citep{wen2020empirical, keskar2017largebatch, chaudhari2017entropysgd, jastrzebski2018factors}.
For example, when evaluated at a global or local minimum, the sharpness metric from \citet{keskar2017largebatch} is equivalent to a scaled version of \mhe, provided that the training objective is well-approximated by its second-order Taylor approximation \citep{keskar2017largebatch}.
\citet{lewkowycz2020large} argued that SGD with large learning rates generalizes well because dynamical instability at initialization causes the optimizer to ``catapult'' to regions where the maximum eigenvalue of the neural tangent kernel and of the Hessian are smaller.
\citet{foret2020sharpness} proposed Sharpness-Aware Minimization (SAM), 
aiming to improve generalization by \emph{explicitly} penalizing a sharpness metric,
and has grown in popularity 
\citep{DBLP:journals/corr/abs-2106-01548, DBLP:journals/corr/abs-2202-00661}.
While the final SAM procedure (their Algorithm 1) does not directly penalize \mhe, the idealized regularizer in their Equation 1 is, at a local or global minimum, equivalent to the maximum Hessian eigenvalue, if one assumes that the training objective is well-approximated by its second-order Taylor approximation.
When \emph{not} at a local or global minimum, the maximum Hessian eigenvalue can be used to drive an upper bound on the idealized SAM regularizer (see Appendix \ref{appendix:a}).

\textbf{Known issues with generalization metrics based on flatness \enspace}
\citet{dinh2017sharp} identified the following issues with sharpness-based metrics.
Given a ReLU network,
one can rescale the weights 
without affecting generalization
(or even the function)
\citep{dinh2017sharp};
this rescaling takes advantage of the fact that
for e.g. a one hidden layer ReLU network with parameters $(\theta_1, \theta_2)$, 
$\{(\alpha\theta_1, \alpha^{-1}\theta_2), \alpha > 0\}$
is an infinite set of observationally equivalent parameters with arbitrarily high sharpness as $\alpha \to \infty$.
Thus,
\citet{dinh2017sharp} theoretically showed that large \mhe\
(i.e., sharp minima)
are not necessarily bad for generalization.
By the same token,
\citet{granziol2020flatness} empirically demonstrated that L2 regularization
can lead to minima with larger \mhe, which generalize well.
\citet{https://doi.org/10.48550/arxiv.2103.06219} showed that the correlation between the Hessian spectral norm 
(which is closely related to \mhe)
and generalization
breaks down when switching the optimizer from 
SGD to Adam.
We now discuss how our findings differ from the aforementioned 
works.
First, the counterexample in \citep{dinh2017sharp} is artificial: these reparameterized models are unnatural and would likely not be reached by SGD. 
Given the empirical nature of our experiments, 
our flat minima are reachable by SGD.
Moreover, 
we do not manipulate the model parameters
and still show
limitations to the connection between
\mhe\
and generalization.
Next,
unlike the negative examples in 
\citep{granziol2020flatness, https://doi.org/10.48550/arxiv.2103.06219},
our settings are simple (i.e., no learning rate schedules)
and hold when SGD is the optimizer. 
And, we provide \emph{multiple} ways to control \mhe, beyond L2 regularization \citep{granziol2020flatness}.
Given the simplicity of our experiments and the number of settings we provide,
our work demonstrates that 
limitations to the connection between
\mhe\
and generalization
are widespread,
rather than subject to
few caveats \citep{dinh2017sharp, granziol2020flatness, https://doi.org/10.48550/arxiv.2103.06219}.
As such, our work highlights the importance of understanding the relationship between \mhe\ and generalization,
and
questions whether 
\mhe\
should be treated as a generalization metric
at all.

\textbf{Large learning rates lead to smaller \mhe \enspace}
Several prior works have observed that training with larger learning rates causes \mhe\ to be smaller along the optimization trajectory \citep{jastrzebski2018factors, jastrzebski2020breakeven, cohen2021gradient}.
In at least the special case of full-batch gradient descent, this relationship is due to dynamical instability \citep{lewkowycz2020large, cohen2021gradient}: gradient descent with learning rate $\eta$ will rapidly escape from any region of the loss landscape where \mhe\ exceeds the value $2/\eta$, and so in practice, gradient descent with learning rate $\eta$ spends most of training in regions of the loss landscape where \mhe\ $\approx 2/\eta$ \citep{cohen2021gradient}.
In the more general case of stochastic gradient descent, the association between large learning rates and small \mhe\ has also been hypothesized to be related to dynamical instability \citep{wu2018how, jastrzebski2020breakeven}.

\textbf{Linear scaling rule \enspace}
The linear scaling rule states that in the small-batch regime, one can maintain generalization performance when increasing batch size
in minibatch SGD by holding fixed the ratio of batch size to learning rate 
\citep{DBLP:journals/corr/GoyalDGNWKTJH17, DBLP:journals/corr/Krizhevsky14, https://doi.org/10.48550/arxiv.1606.04838, jastrzebski2018factors}.
\citet{jastrzebski2018factors}
finds that smaller values of the batch size to learning rate ratio correlate with improved generalization
and small \mhe, 
which is consistent with the 
popular intuition
that small batch sizes
(and large learning rates) generalize better 
by guiding SGD towards ``flatter'' solutions 
\citep{jastrzebski2018factors, smith2018bayesian, lewkowycz2020large}.

\section{Experiments}
\label{sec:experiments}
In this section, we provide 
four examples\footnote{We use the exact same weight initializations (i.e. the same seeds) in each experimental setting.
} 
of training interventions that 
promote smaller values for 
\mhe,
but either do not change, or degrade, generalization.
We then provide one example of an intervention that improves generalization, without changing
\mhe\footnote{Due to compute constraints, we calculate \mhe\ over a subset of the training dataset.
}.
\subsection{Scaling learning rates in the small- and large-batch regimes}
\label{sec:sgd}

\begin{figure}[!ht]
    \centering
    \includegraphics[width=\textwidth]{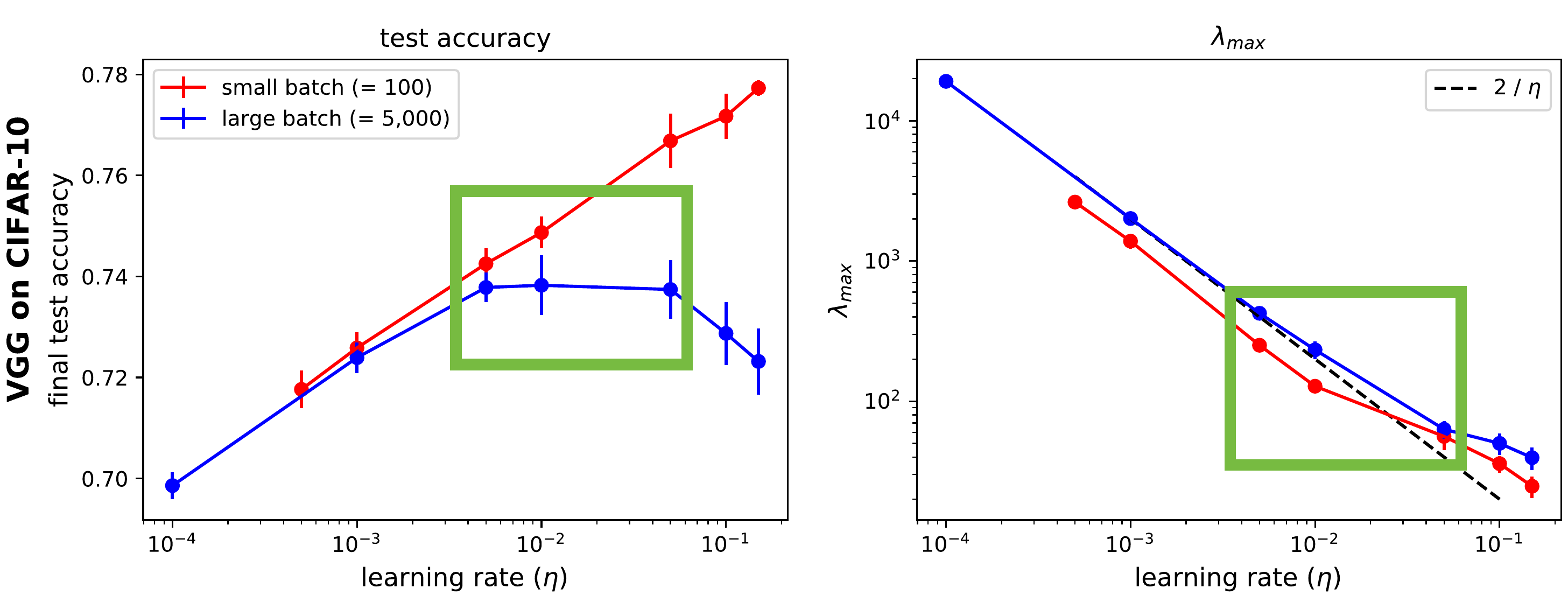}
    \caption{
    \textbf{Small \mhe\ do not always generalize better.}
    We train a VGG11 (no BN) using SGD at a range of learning rates. 
    We plot the final test accuracy and \mhe\ 
    (averaged over 5 runs with different random seeds) 
    as a function of the learning rate. 
    Observe that as the learning rate increases,
    the final test accuracy increases for small-batch 
    and plateaus for large-batch SGD. 
    In both cases, as the learning rate increases,
    \mhe\ decreases.}
    \label{fig:vgg-main}
\end{figure}

We demonstrate that when a VGG-11 without batch normalization
is trained on CIFAR-10 in the large-batch setting, 
larger learning rates cause \mhe\ to be smaller, 
yet do not promote better generalization.
(By contrast, in the \emph{small-batch} setting, 
larger learning rates both  
reduce \mhe\ and improve generalization.)
For the plots in Figure \ref{fig:vgg-main}, 
we train with minibatch SGD 
until reaching 99\% train accuracy. 
We consider various learning rates and both the small-batch (red) and large-batch (blue) regime. 
In our experiments, we use 
small batch size 100 and 
large batch size 5000.
For \emph{small-batch SGD}, we make the following observations.

\textbf{(i) For small-batch SGD, large learning rates generalize better \enspace}
We plot the test accuracy attained 
by small-batch SGD at varying learning rates
(Figure \ref{fig:vgg-main}).
Observe that
small-batch SGD generalizes better 
when the learning rate is larger,
as observed previously
\citep{jastrzebski2018factors, NEURIPS2019_dc6a7071}.

\textbf{(ii) For small-batch SGD, large learning rates find small \mhe \enspace}
We plot \mhe\ attained 
by small-batch SGD at varying learning rates
and
observe that \mhe\ is smaller 
when the learning rate is large 
(Figure \ref{fig:vgg-main}).
These findings also have precedents in the literature \citep{jastrzebski2018factors, NEURIPS2019_dc6a7071}.

Moreover, these two findings are consistent with the popular hypothesis 
that a small \mhe\ is the \textit{causal} mechanism 
by which large learning rates promote generalization.
However, while this hypothesis is consistent with experimental data for \emph{small-batch} SGD, 
we will see that it is \emph{not} consistent with experimental data for large-batch or full-batch SGD.

\textbf{(iii) For large-batch SGD, 
large learning rates still find small \mhe \enspace}
As with \emph{small-batch} SGD, large learning rates also find solutions with small \mhe\
for \emph{large-batch} SGD
(Figure \ref{fig:vgg-main}).
The mechanism behind this phenomenon is well-understood: as \cite{cohen2021gradient} showed, full-batch gradient descent typically operates in a regime (the Edge of Stability),
in which \mhe\ constantly hovers 
just above the numerical value $\frac{2}{\eta}$,
where $\eta$ is the learning rate.
Thus, for large-batch SGD, \mhe\ at the gradient descent solution is approximately $\frac{2}{\eta}$.
To confirm this, we draw a dotted black line
to mark the line  $\frac{2}{\eta}$; 
we can see that \mhe\ of the large-batch SGD solution
lies approximately on the line.

\textbf{(iv) But, for large-batch SGD, large learning rates don't generalize better \enspace}
Beyond some point, large learning rates \textbf{do not} generalize better than small learning rates for large-batch SGD (Figure \ref{fig:vgg-main}).
Thus, in this experiment, smaller \mhe\ is not always correlated with better generalization.

\begin{figure}[!ht]
    \centering
    \includegraphics[width=\textwidth]{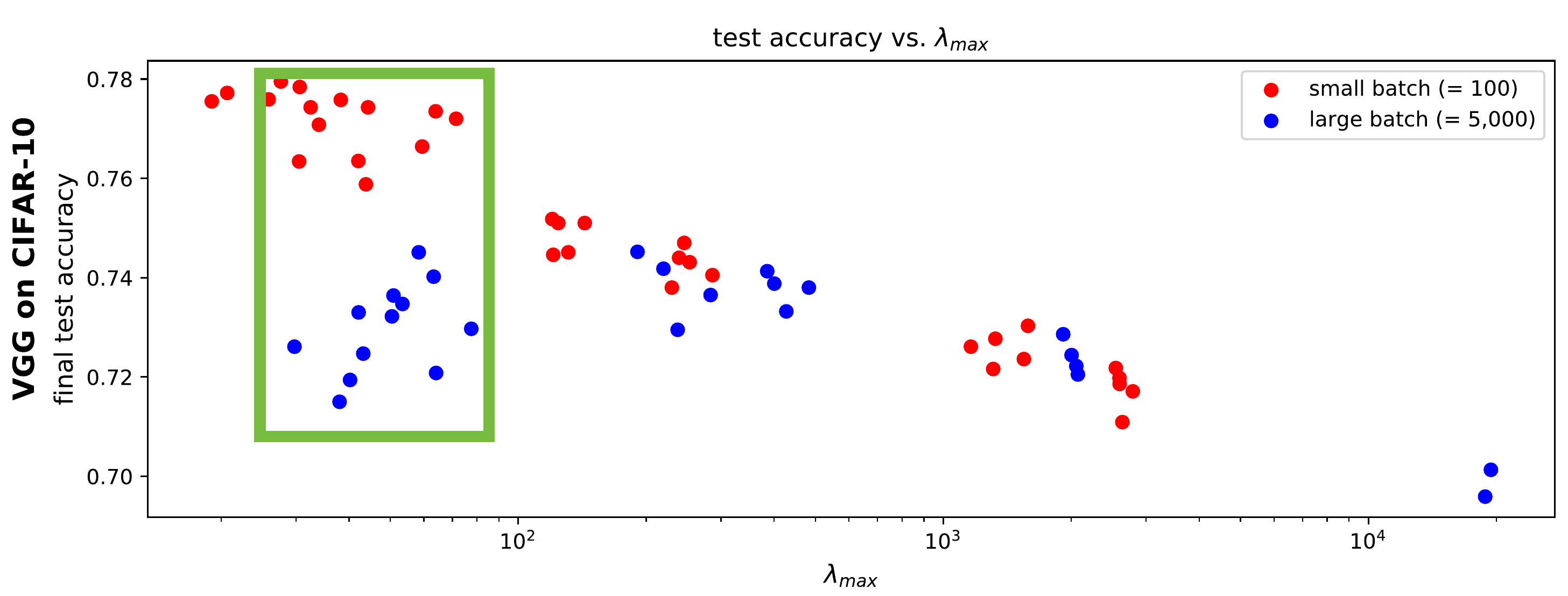}
    \caption{
    \textbf{Equivalent \mhe\ can generalize differently}
    We train a VGG11 (no BN) using SGD at a range of learning rates. We present a scatter plot of \mhe\ and test accuracy for both small and large-batch SGD. Observe that in the green box, there are multiple instances where \mhe\ of small-batch and large-batch training runs are similar, but the test accuracy is much better for small-batch SGD.}
    \label{fig:vgg-main-scatter}
\end{figure}

\textbf{No absolute correlation between \mhe\ and test accuracy \enspace}
Observe that \mhe\ at small batch sizes is lower than \mhe\ at large-batch size 
(Figure \ref{fig:vgg-main}).
Thus, one might attempt to resuscitate the conventional wisdom 
by hypothesizing that 
a small \mhe\ is
indeed correlated with good generalization,
but that this effect only ``kicks in'' once \mhe\ is sufficiently low.
However, this is not the case:
observe that there are multiple regions 
in Figure \ref{fig:vgg-main-scatter}
where \mhe\ of small and large-batch training runs are similar, but the test accuracy is much better for small-batch SGD.
Thus, even ultra-small values of \mhe\ are not always associated with improved generalization.

\begin{figure}[!ht]
    \centering
    \includegraphics[width=\textwidth]{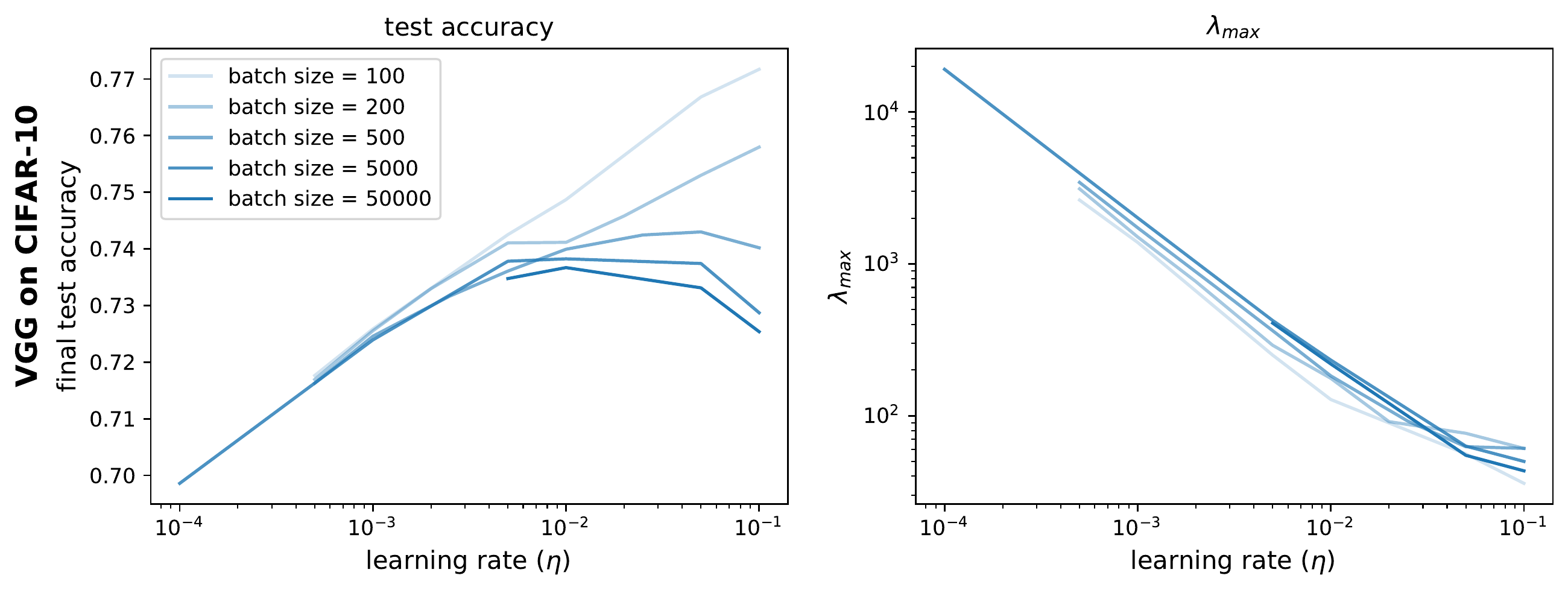}
    \caption{
    \textbf{Generalization benefits of large learning rates diminish (and sometimes reverse) as batch size increases}
    We show our findings from Figure \ref{fig:vgg-main} hold for a large range of batch sizes. Observe that as the batch size gets larger, the generalization benefits of large learning rates gradually diminish, yet large learning rates still find 
    small \mhe.}
    \label{fig:vgg-gradual}
\end{figure}
\vspace{10px}
\textbf{A gradual trend \enspace}
Previously, we presented data for only two batch sizes: 
$100$ and $5000$ (Figure \ref{fig:vgg-main}).
To fill in the picture, we present results
for a range of batch sizes (Figure \ref{fig:vgg-gradual}).
Note that the effect we previously observed is gradual.
As the batch size gets larger, 
the generalization benefits 
of large learning rates gradually diminish, 
yet large learning rates still find 
solutions with small \mhe.
As such, we show that $\lambda_{max}$ 
is not a robust generalization measure 
in the case of large-batch/full-batch gradient descent, 
which is in line with 
the work in
\citep{DBLP:journals/corr/abs-2010-11924}
on the success of a theory (in this case, of a generalization measure) for a family of environments.
\subsection{When scaling learning rate and batch size simultaneously in the small-batch regime, \mhe\ and generalization aren't correlated}\label{lsr}

According to the linear scaling rule, 
in the small-batch regime, models trained 
with the same batch size to learning rate ratio 
(denoted by $\alpha$) 
should exhibit similar generalization behavior \citep{DBLP:journals/corr/GoyalDGNWKTJH17, DBLP:journals/corr/Krizhevsky14, https://doi.org/10.48550/arxiv.1606.04838, jastrzebski2018factors}.
Meanwhile, prior work suggests that models trained with large learning rates should have small \mhe\ \cite{jastrzebski2018factors, jastrzebski2020breakeven}.
Consequently, one might suspect that in the small-batch regime, if we scale learning rate and batch size simultaneously while keeping their ratio fixed, \mhe\ will get smaller while generalization will remain approximately unchanged---providing another counterexample in which generalization and \mhe\ are not correlated.
We now confirm that this suspicion is true.

\begin{figure}[!ht]
    \centering
    \includegraphics[width=\textwidth]{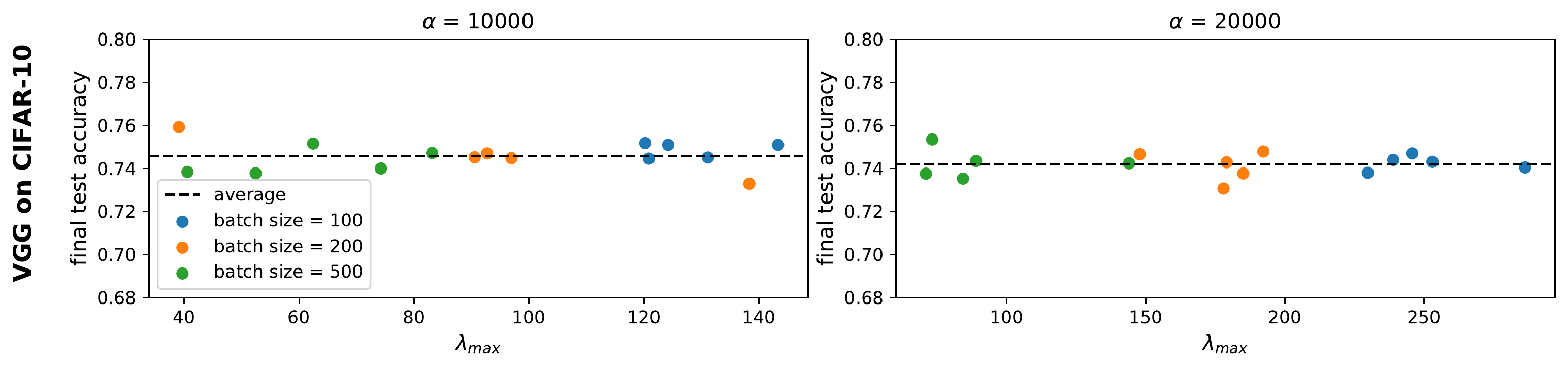}
    \caption{
    \textbf{Test accuracy, but not $\mathbf{\lambda_{max}}$, depends on ratio of batch size to learning rate}
    Let $\alpha$ denote the ratio of batch size to learning rate.
    We plot the final test accuracy vs. \mhe\ for various $\alpha$, with 5 runs for every (batch size, $\eta$) pair. Observe that networks trained with the same $\alpha$ but different batch sizes exhibit similar final test accuracies, but different \mhe.
    }
    \label{fig:vgg-lsr}
    \vspace{-8px}
\end{figure}

\begin{figure}[!ht]
    \centering
    \includegraphics[width=\textwidth]{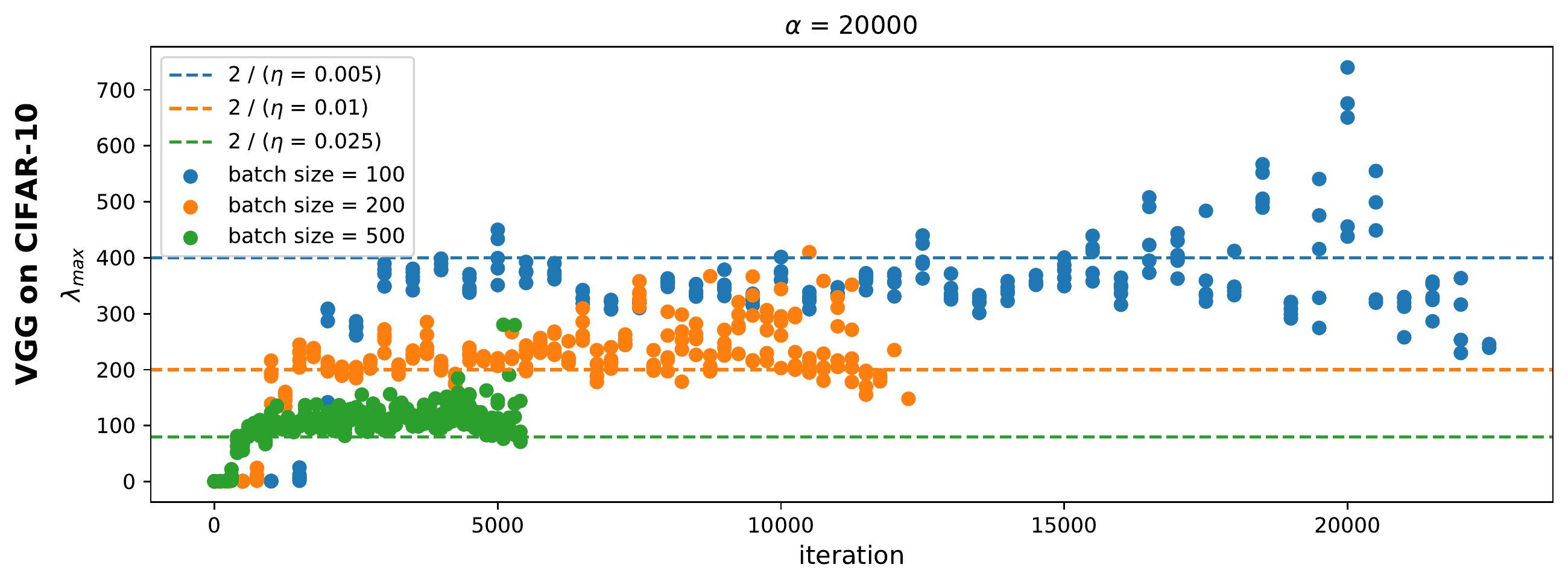}
    \caption{
    \textbf{For a fixed batch size to learning rate ratio, larger learning rates find smaller \mhe}
    For $\alpha$ = 20000, we plot the evolution of \mhe\ during training over 5 runs for every (batch size, $\eta$) pair.
    Observe that \mhe\ tends to remain the smallest for $\eta$ = 0.025 and largest for $\eta$ = 0.005.
    }
    \label{fig:vgg-lsr-evolution}
    \vspace{-10px}
\end{figure}

We confirm the linear scaling rule with respect to final test accuracy, but find that \mhe\ behaves differently (Figure \ref{fig:vgg-lsr}).
In particular, for a fixed $\alpha$, 
observe that \mhe\ is large when the batch size (and corresponding learning rate) is small.
We attribute this phenomenon to the influence of learning rate on \mhe.
In Figure \ref{fig:vgg-lsr-evolution}, we plot the evolution of \mhe\ for a fixed $\alpha$, using three different (batch size, $\eta$) pairs.
We observe that larger learning rates lead to smaller \mhe\ throughout training.
This is consistent with our observation in Figure \ref{fig:vgg-main}, where larger learning rates lead to smaller \mhe, regardless of the batch size.

\subsection{Sharpness-Aware Minimization can degrade generalization for large batch sizes}\label{SAM}

Motivated by the intuition that flatness causally promotes generalization, \citet{foret2020sharpness} proposed an algorithm, 
Sharpness-Aware Minimization (SAM), 
which aims to improve generalization by directly penalizing a sharpness metric.
Their idealized sharpness metric, given by their Equation 1, is the worst-case loss within a ball, $\max_{\| \boldsymbol{\epsilon} \|_2 \le \rho} L(\mathbf{w} + \boldsymbol{\epsilon})$.
If one assumes that $L$ is locally well-approximated by its quadratic Taylor approximation, and if one assumes that $\mathbf{w}$ is a local minimizer (so that the gradient vanishes), then this idealized sharpness metric is equivalent to 
$L(\mathbf{w}) + \frac{\rho^2}{2} \, \lambda_{\max}(\mathbf{w})$.
In the more general situation where we make the quadratic Taylor approximation but $\mathbf{w}$ is not a local minimum, the idealized sharpness metric is upper-bounded by 
$L(\mathbf{w}) + \rho \|\nabla L(\mathbf{w})\| + \frac{\rho^2}{2} \lambda_{\max}(\mathbf{w})$ 
(see Appendix \ref{appendix:a}).
In practice, \citet{foret2020sharpness} do not propose to penalize their idealized metric; instead, they propose to penalize a proxy for 
$\rho \| \nabla L(\mathbf{w}) \|$,
the first-order version of their idealized metric.

\begin{figure}[ht]
    \centering
    \includegraphics[width=\textwidth]{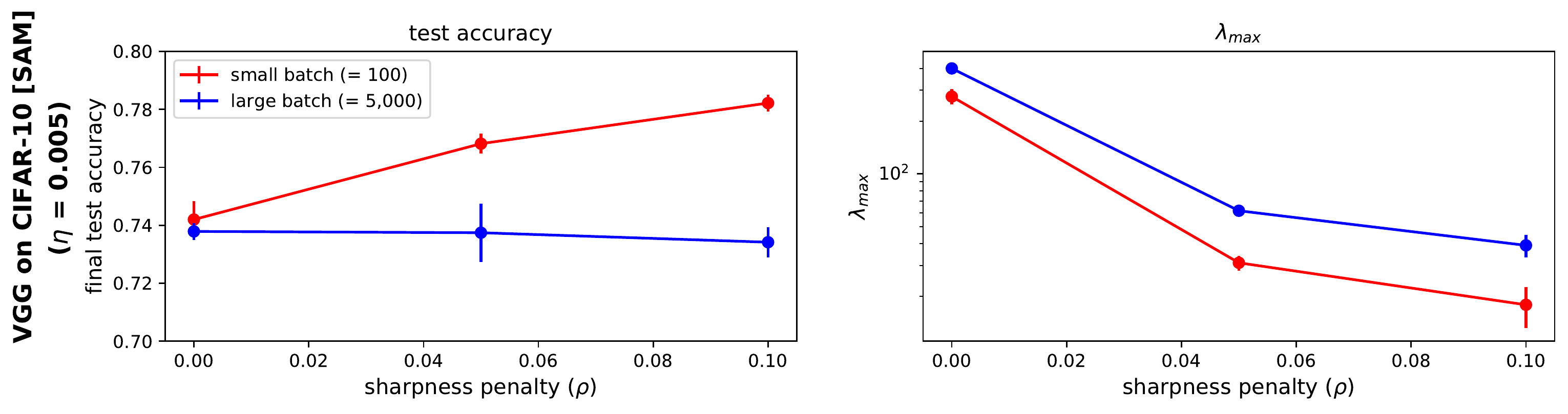}
    \caption{
    \textbf{SAM only exhibits generalization benefits for small batch sizes.}
    We train a VGG11 (no BN) using SAM until achieving 99\% train accuracy at a range of values for $\rho$ with a learning rate of 0.005. We plot the final test accuracy and \mhe\ (averaged over 4 runs with different random seeds) as a function of $\rho$. Observe that as $\rho$ increases, the final test accuracy increases for small batch and decreases (slightly) for large batch. In both cases, as $\rho$ increases, \mhe\ decreases.}
    \label{fig:vgg-sam-1}
\end{figure}

\begin{figure}[!ht]
    \centering
    \includegraphics[width=\textwidth]{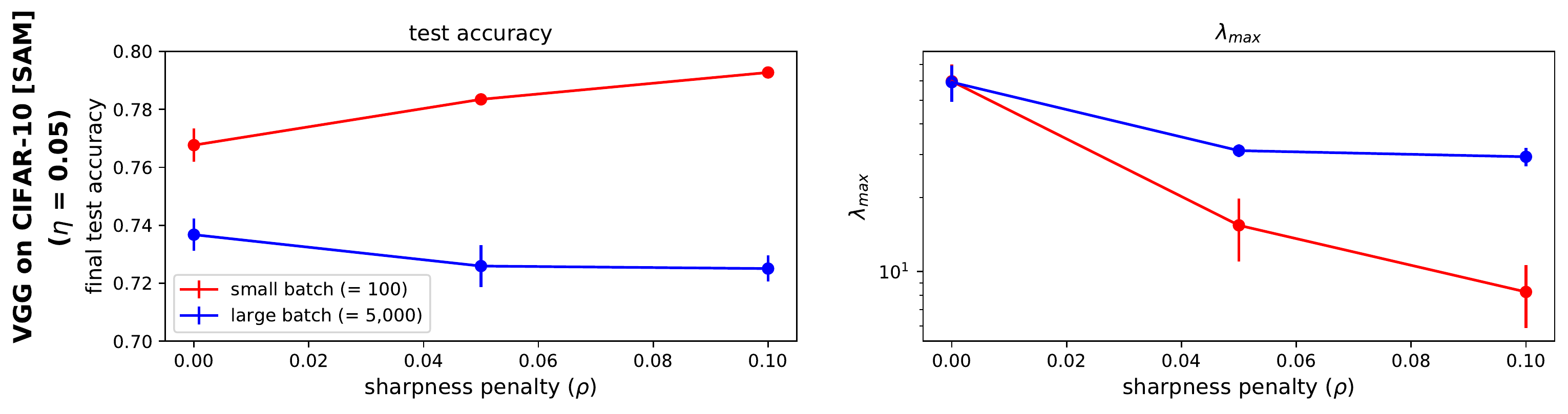}
    \caption{
    We use the same setup as in Figure \ref{fig:vgg-sam-1}, but with the large learning rate 0.05. The results are consistent with Figure \ref{fig:vgg-sam-1}.
    }
    \label{fig:vgg-sam-2}
\end{figure}

The SAM algorithm has a hyperparameter $\rho$ which controls the strength of the sharpness penalty.
We use SAM\footnote{We use David Samuel's PyTorch implementation of SAM: \url{https://github.com/davda54/sam}.} 
to train non-normalized VGG networks on CIFAR10 with cross-entropy loss until we achieve 99\% train accuracy. 
We consider $\eta=0.005$ (Figure \ref{fig:vgg-sam-1})  and $\eta=0.05$
(Figure \ref{fig:vgg-sam-2}) in both the small-batch (red lines) and large-batch (blue lines) regime. At small batch sizes, SAM works as intended: 
in Figures \ref{fig:vgg-sam-1} and \ref{fig:vgg-sam-2}, in red, we plot both the test accuracy (left column) and \mhe\ (right column) as we vary $\rho$,
the SAM hyperparameter. 
Observe that for the small batch size,
a higher $\rho$ (higher sharpness penalty) causes the test accuracy to be higher, 
and \mhe\ to be lower.

However,
at large batch sizes, SAM
does not work as intended.
Plotting the test accuracy and \mhe,
we find that for large batch size, 
higher $\rho$ (sharpness penalty) 
still causes \mhe\ to shrink, 
but this no longer improves test accuracy.
(This is in line with the observation in \citep{https://doi.org/10.48550/arxiv.2206.06232}
that increasing $\rho$ does not always improve generalization
in the large-batch regime.)
Indeed, at the large batch size, a higher $\rho$ 
can lower the final test accuracy,
even as it finds 
smaller \mhe.

\vspace{-5px}
\subsection{Dropout in DNNs}
\begin{figure}[!ht]
    \centering
    \includegraphics[width=\textwidth]{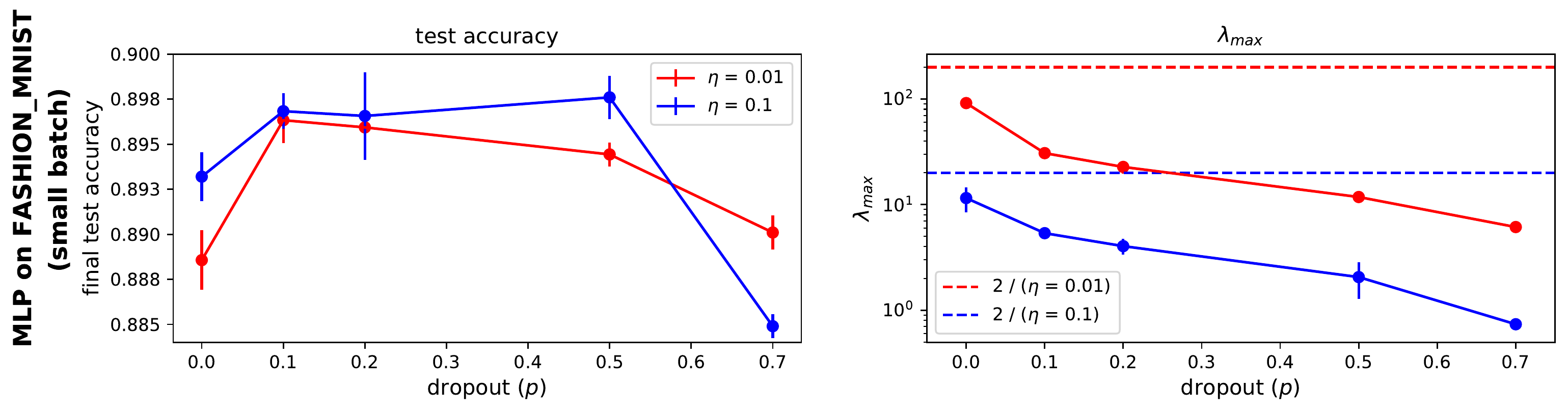}
    \caption{
    \textbf{Dropout induces small \mhe, yet does not always boost generalization.}
    We train an MLP (with 2 hidden layers and dropout) using small-batch SGD at a range of learning rates. We plot the final test accuracy and \mhe\ (averaged over 3 runs with different random seeds) as a function of the dropout probability $p$. Observe that models with higher dropout probabilities (until $p=0.2$) exhibit slight generalization benefits, and that higher values of $p$ lead to lower \mhe.}
    \label{fig:mlp-dropout-sb}
\end{figure}

\begin{figure}[!ht]
    \centering
    \includegraphics[width=\textwidth]{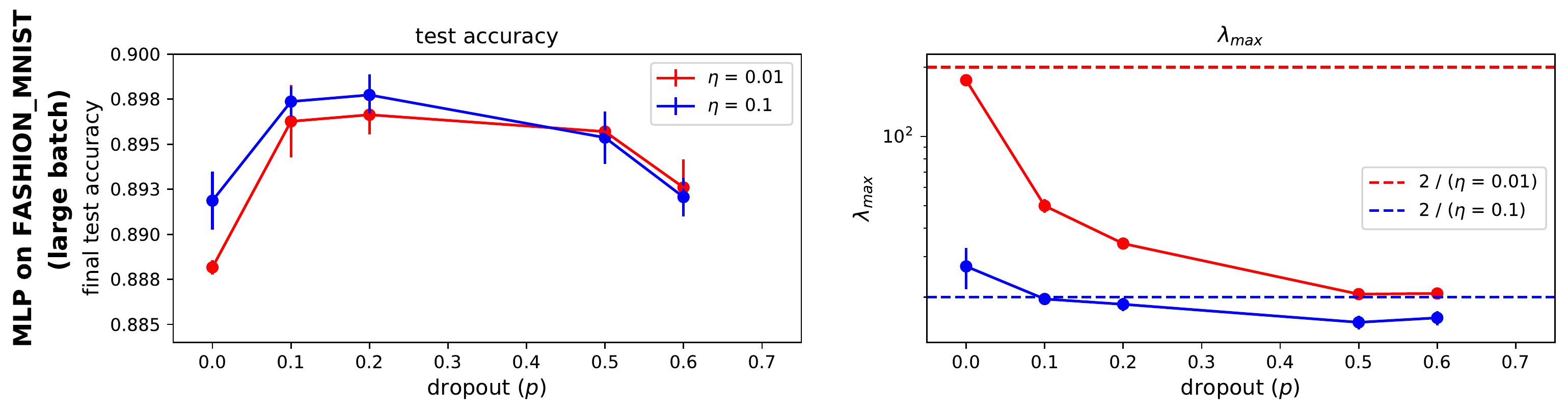}
    \caption{
    We use the same setup as in Figure \ref{fig:mlp-dropout-sb} with a larger batch size. Our observations are consistent with those in Figure \ref{fig:mlp-dropout-sb}.}
    \label{fig:mlp-dropout-lb}
\end{figure}

We now demonstrate that while higher dropout  \citep{JMLR:v15:srivastava14a} generally causes \mhe\ to shrink,
but that at excessive dropout levels,
these \mhe\ are accompanied
by degradations in generalization
(Figures \ref{fig:mlp-dropout-sb} and \ref{fig:mlp-dropout-lb}).
In these experiments, we train networks on Fashion-MNIST 
with cross-entropy loss and minibatch SGD 
until achieving 99\% train accuracy. 
We consider various learning rates, batch sizes, and dropout probabilities $p$.
In these experiments, our model is an MLP with 2 hidden layers, with dropout applied to both.
Concretely, we make the following observations.

\textbf{(i) For a fixed learning rate and batch size, modest dropout rates improve generalization \enspace}
We plot the test accuracy attained by the model with dropout probabilities $p$ at varying learning rates for both small-batch and large-batch (Figures \ref{fig:mlp-dropout-sb} and \ref{fig:mlp-dropout-lb}).
Observe that for both small-batch and large-batch SGD, adding dropout (up to $p=0.2$) improves generalization.

\textbf{(ii) High dropout probabilities promote small \mhe \enspace}
Higher dropout probabilities induce flatter solutions, 
as measured by smaller \mhe\ 
(Figures \ref{fig:mlp-dropout-sb} and \ref{fig:mlp-dropout-lb}).
This observation holds true in both the small-batch and large-batch regime.
Note that we turn off dropout when computing \mhe\ because that is how predictions are generally made.

Recall that for a fixed learning rate,
\mhe\ at small batch sizes 
is lower than \mhe\ at large batch sizes 
(see Figure \ref{fig:vgg-main} 
and \citep{jastrzebski2018factors, jastrzebski2020breakeven}).
We hypothesize that randomness due to high dropout rates and stochasticity introduced by smaller batch sizes promote smaller \mhe\ by a similar mechanism.

\textbf{(iii) However, excessively high dropout probabilities don't generalize better \enspace}
For each learning rate in both 
the small-batch and large-batch regimes,
\mhe\ is lower for higher dropout rates.
However, as the dropout probabilities 
are made \emph{excessively} high,
lower \mhe\ is accompanied by worse generalization. 
For example, for batch size 100 and $\eta=0.1$,
the solution found using $p=0.7$ has a smaller \mhe\ than the one found using $p=0.1$, 
yet generalizes worse (Figure \ref{fig:mlp-dropout-sb}).
\subsection{Batch Normalization in DNNs}
In each previous example, the interventions
reduced \mhe\ but
either did not change,
or even degraded,
generalization.
We now demonstrate that adding Batch Normalization (BN) has the opposite character: it leaves \mhe\ unchanged, but alters generalization
(Figures \ref{fig:vgg-bn-sb} and \ref{fig:vgg-bn-lb}).
In these experiments,
we train VGG networks both with and without BN.
As before,
we train using minibatch SGD on CIFAR10 until achieving 99\% train accuracy.
We observe the following:

\begin{figure}[!ht]
    \centering
    \includegraphics[width=\textwidth]{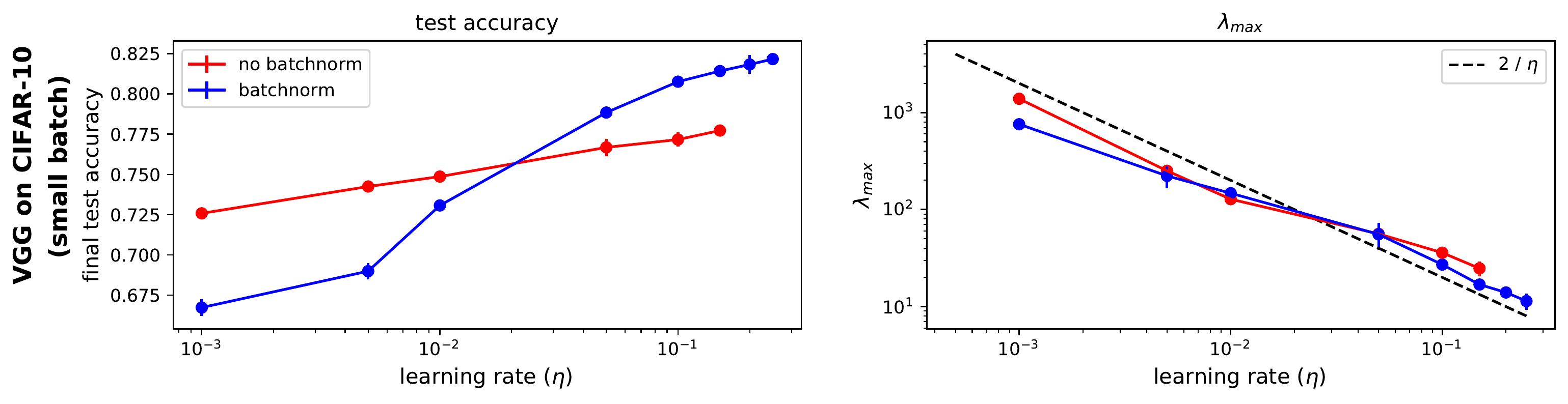}
    \caption{
    \textbf{BN can impact generalization without significantly affecting \mhe}
    We train a VGG11 (+ BN) and VGG11 (no BN) using small-batch SGD at a range of learning rates. We plot the final test accuracy and \mhe\ (averaged over 4 runs with different random seeds) as a function of the learning rate. 
    Observe that for a fixed learning rate, the VGG11 (+BN) and the VGG11 (no BN) exhibit different test accuracies, yet have comparably sharp solutions.
    }
    \label{fig:vgg-bn-sb}
\end{figure}

\begin{figure}[!ht]
    \centering
    \includegraphics[width=\textwidth]{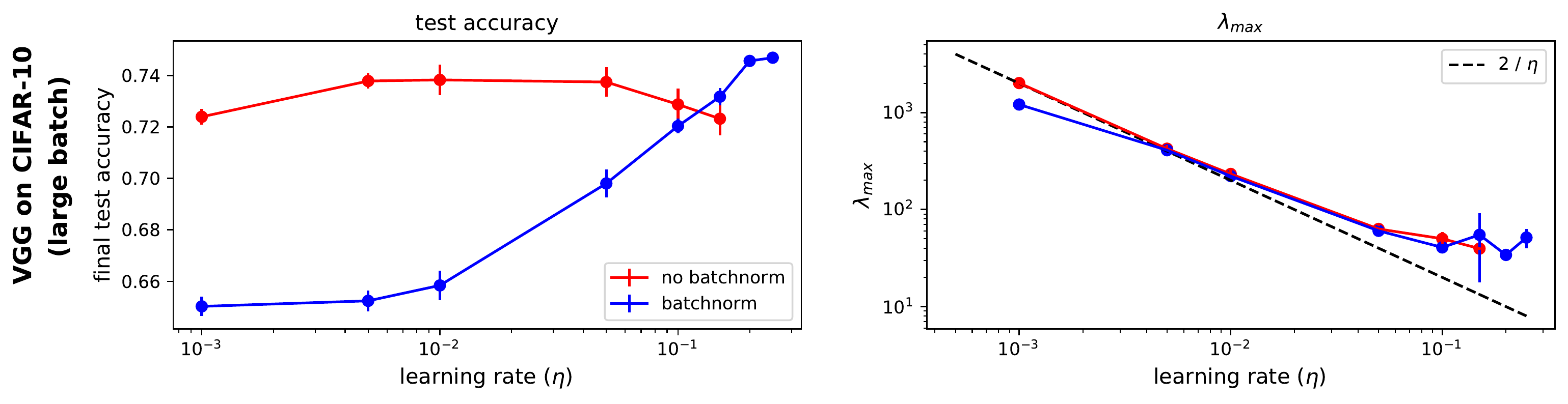}
    \caption{
    We use the same setup as in Figure \ref{fig:vgg-bn-sb} with a large batch size. Our observations are consistent with those in Figure \ref{fig:vgg-bn-sb}.
    }
    \label{fig:vgg-bn-lb}
    \vspace{-15px}
\end{figure}
\vspace{15px}
\textbf{(i) Models with/without BN generalize differently \enspace}
We observe that 
for \emph{small} learning rates, 
models with BN generalize \emph{worse} than models without BN;
but for \emph{large} learning rates, 
models with BN generalize \emph{better} than models without BN
(Figures \ref{fig:vgg-bn-sb} and \ref{fig:vgg-bn-lb}).
Both findings are consistent with \citep{DBLP:journals/corr/abs-1806-02375}.
(While in Section \ref{sec:sgd},
large learning rates did not improve generalization 
for unnormalized networks in the large-batch setting, we find here that for networks with BN enabled, 
large learning rates do improve generalization
in the large-batch setting.)

We will now compare \mhe\ found by models with and without BN.
As noted in \citet{cohen2021gradient},
for BN networks trained with small $\eta$,
\mhe\ should be measured
in between successive iterates, 
rather than directly at the iterates themselves.
To estimate \mhe\ in this regime,
we follow the procedure detailed in \citep{cohen2021gradient},
where we compute the maximum \mhe\ over a grid of eight points spaced evenly
between two successive iterates (specifically, we use the final iterate and train the network for one extra iteration).

\textbf{(ii) However, for fixed learning rates and batch size, \mhe\ is not sensitive to BN \enspace}
We plot \mhe\ attained by SGD and BN at varying learning rates (Figures \ref{fig:vgg-bn-sb} and \ref{fig:vgg-bn-lb}).
Observe that in the large-batch setting (Figure \ref{fig:vgg-bn-lb}), if we fix the learning rate, then \mhe\ found by models with and without BN are equivalent for most values of $\eta$. 
This is consistent with \citep{cohen2021gradient}, which demonstrates that networks trained via full-batch gradient descent still train at the edge of stability in the presence of BN.
Additionally, this finding provides more evidence 
in support of the idea that
solutions with equivalent \mhe\
can generalize differently, as we observe in Figure \ref{fig:vgg-main-scatter}.
(Note that 
$\eta=0.001$,
\mhe\ found for the model with BN is slightly smaller than without BN. We suspect that \mhe\ does not match up exactly because we use a batch size of 5000, rather than the full batch; in general, smaller minibatches result in lower \mhe\ \citep{jastrzebski2020breakeven}.)
Likewise, in the small-batch setting (Figure \ref{fig:vgg-bn-sb}), if we fix $\eta$, then 
\mhe\ found by models with and without BN are highly similar for most $\eta$,
despite large differences in test accuracy.
We acknowledge that in the small-batch setting, 
\mhe\ of un-normalized networks is slightly higher at large learning rates,
and \mhe\ of normalized networks is lower at small learning rates.
Importantly, in the latter case, normalized networks exhibit lower test accuracies,
which is the \emph{opposite}
of what one might expect given the popular intuition that 
small \mhe\
cause good generalization.

\section{Conclusion}
\label{sec:conclusion}
We have demonstrated that by manipulating the learning rate, batch size, training algorithm, and dropout, we can decrease the $\lambda_{\max}$ of a solution; yet these solutions with small $\lambda_{\max}$ do not always generalize better.
Similarly, we demonstrated that we can boost generalization performance without promoting 
small \mhe,
e.g., by using batch normalization.
Our findings add to the growing body of work that calls into question the influence of
flatness on generalization \citep{dinh2017sharp}.
Since this paper refers to flatness as \mhe\ (the leading Hessian eigenvalue),
note that our findings do not necessarily hold for other flatness metrics (e.g., the Hessian trace).
While methods motivated by flatness produce useful tools \citep{foret2020sharpness},
\mhe\
does not provide a scientific explanation for improvements in generalization.
Thus, it is evident that there is a deeper story behind
why flatness seems to be fruitful intuition.
We hope to inspire future efforts aimed at understanding the relationship between \mhe\ and generalization, and to 
facilitate further discussion
regarding whether \mhe\ should be treated as a generalization metric at all.


\bibliographystyle{unsrtnat}
\bibliography{refs}

\appendix
\section{Appendix}
\subsection{An Upper Bound on the original Sharpness Aware Minimization (SAM) Objective}\label{appendix:a}
Below, we will show that 
one way to penalize an upper bound on the original 
Sharpness Aware Minimization
objective would be to penalize the 
leading Hessian eigenvalue, \mhe. 

\citet{foret2020sharpness} define the Sharpness Aware Minimization problem as follows:
\begin{equation}
    \min_{\mathbf{w}} \max_{\|\boldsymbol{\epsilon}\|_2 \leq \rho} L(\mathbf{w} + \boldsymbol{\epsilon}) + \gamma\|\mathbf{w}\|^2_2 
\end{equation}

where $\rho$ and $\gamma$ are hyperparameters.
Using a second-order Taylor approximation, we can see that
\begin{align*}
    L(\mathbf{w} + \boldsymbol{\epsilon}) 
    &\approx  
    L(\mathbf{w}) + \nabla L(\mathbf{w})^\top\boldsymbol{\epsilon} + \frac{1}{2}\boldsymbol{\epsilon}^\top (\nabla^2 L(\mathbf{w})) \boldsymbol{\epsilon} \\
\end{align*}
Given some $\boldsymbol{\epsilon}$ with $\|\boldsymbol{\epsilon}\| \leq \rho$, 
$\exists ~ \hat{\boldsymbol{\epsilon}}=\frac{\boldsymbol{\epsilon}}{\|\boldsymbol{\epsilon}\|}$.
Since $\hat{\boldsymbol{\epsilon}}^\top (\nabla^2 L(\mathbf{w})) \hat{\boldsymbol{\epsilon}} \leq \lambda_{\max}$,
$\boldsymbol{\epsilon}^\top (\nabla^2 L(\mathbf{w})) \boldsymbol{\epsilon} \leq \rho^2 \lambda_{\max}$.

Thus, observe that the following upper bound on the original SAM objective
explicitly regularizes \mhe.
\begin{align*}
    \max_{\|\boldsymbol{\epsilon}\|_2 \leq \rho} L(\mathbf{w} + \boldsymbol{\epsilon}) 
    &\approx  
    \max_{\|\boldsymbol{\epsilon}\|_2 \leq \rho} L(\mathbf{w}) + \nabla L(\mathbf{w})^\top\boldsymbol{\epsilon} + \frac{1}{2}\boldsymbol{\epsilon}^\top (\nabla^2 L(\mathbf{w})) \boldsymbol{\epsilon} \\
    &\leq L(\mathbf{w}) + 
        \left[\max_{\|\boldsymbol{\epsilon}\|_2 \leq \rho} \nabla L(\mathbf{w})^\top\boldsymbol{\epsilon} \right] +
        \left[\max_{\|\boldsymbol{\epsilon}\|_2 \leq \rho} \frac{1}{2}\boldsymbol{\epsilon}^\top (\nabla^2 L(\mathbf{w})) \boldsymbol{\epsilon} \right] \\
    &\leq L(\mathbf{w}) + 
        \left[\max_{\|\boldsymbol{\epsilon}\|_2 \leq \rho} \|\nabla L(\mathbf{w})\|\|\boldsymbol{\epsilon}\| \right] +
        \left[\max_{\|\boldsymbol{\epsilon}\|_2 \leq \rho} \frac{1}{2}(\rho^2 \lambda_{\max}) \right] \\
    &\leq L(\mathbf{w}) + 
        \rho \|\nabla L(\mathbf{w})\| +
        \frac{\rho^2}{2} \lambda_{\max}
        \\
\end{align*}

In this sense, we can think of the hyperparameter $\rho$ as 
the strength of the penalty imposed on sharpness.

Note that in practice, due to computational complexity, \citet{foret2020sharpness} use a first order Taylor approximation for the inner maximization problem and drop second order terms in the gradient approximation of the modified loss function.

\clearpage
\subsection{Linear Scaling Rule (continued)}\label{appendix:b}
\begin{figure}[!ht]
    \centering
    \includegraphics[width=\textwidth]{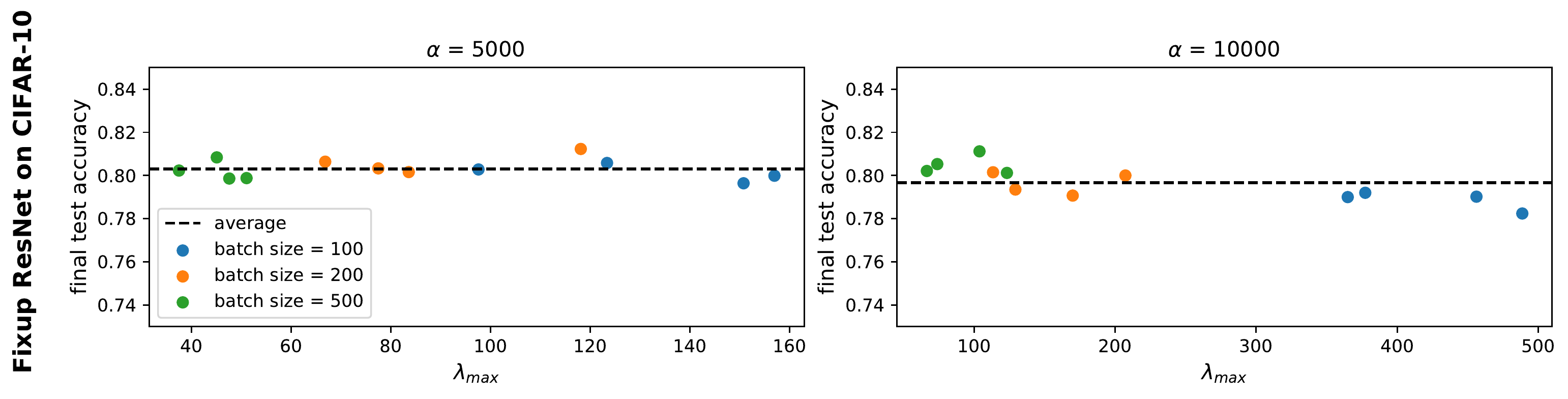}
    \caption{
    \textbf{Test accuracy, but not $\mathbf{\lambda_{max}}$, depends on ratio of batch size to learning rate}
    Let $\alpha$ denote the ratio of batch size to learning rate.
    We plot the final test accuracy vs. \mhe\ for various $\alpha$, with 4 runs for every (batch size, $\eta$) pair. Observe that networks trained with the same $\alpha$ but different batch sizes exhibit similar final test accuracies, but different \mhe.
    These findings are consistent with those in Figure \ref{fig:vgg-lsr}.
    }
    \label{fig:fixup-lsr}
    \vspace{-8px}
\end{figure}

We confirm that our findings from Section \ref{lsr} hold for a Fixup ResNet32 (see Figure \ref{fig:fixup-lsr}).

\clearpage
\subsection{SAM (continued)}\label{appendix:c}
For some architectures, large learning rates and SAM improve the final test accuracy in the large-batch regime, but for other architectures, they do not. 

\begin{figure}[!ht]
    \centering
    \includegraphics[width=\textwidth]{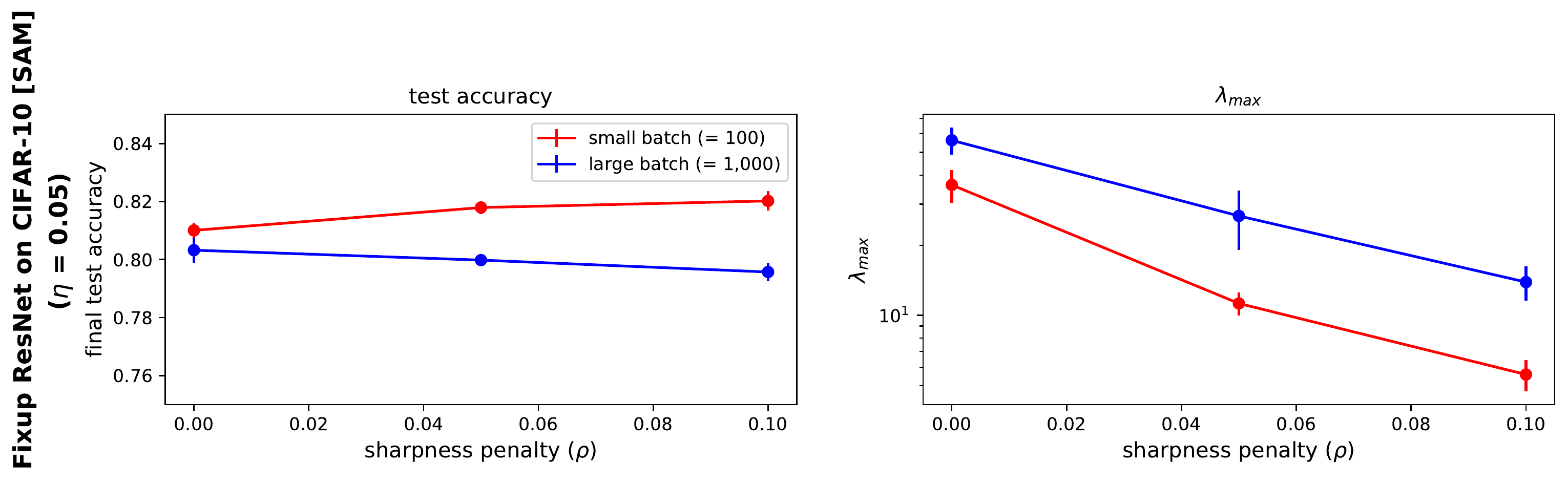}
    \includegraphics[width=\textwidth]{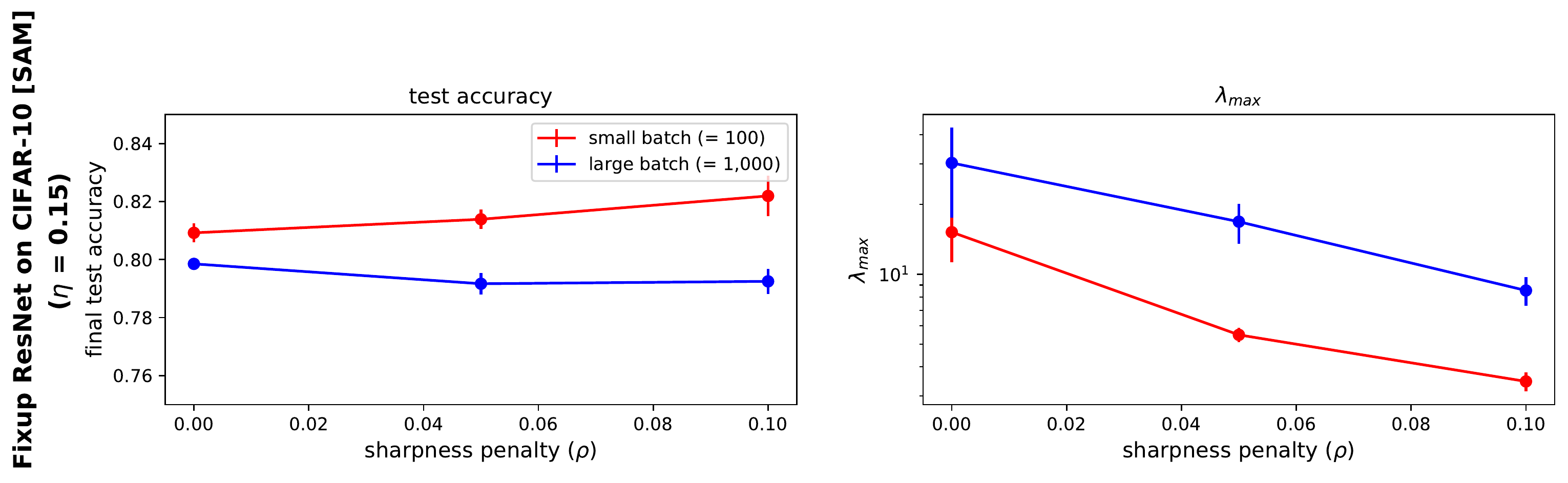}
    \caption{
    \textbf{For networks without BN, SAM only exhibits generalization benefits for small batch sizes.}
    We train a Fixup ResNet32 (no BN) using SAM until achieving 99\% train accuracy at a range of values for $\rho$ with learning rates of 0.05 and 0.15. We plot the final test accuracy and \mhe\ (averaged over 4 runs with different random seeds) as a function of $\rho$. Observe that as $\rho$ increases, the final test accuracy increases for small batch and decreases (slightly) for large batch. In both cases, as $\rho$ increases, \mhe\ decreases.
    These findings are consistent with those in Figures \ref{fig:vgg-sam-1} and \ref{fig:vgg-sam-2}.}
    \label{fig:fixup-sam-large}
    \vspace{-8px}
\end{figure}

We confirm that for a Fixup ResNet32, 
SAM only exhibits generalization benefits for small batch sizes (see Figure \ref{fig:fixup-sam-large}).
This is consistent with our findings in Section \ref{SAM}.

\begin{figure}[!ht]
    \centering
    \includegraphics[width=\textwidth]{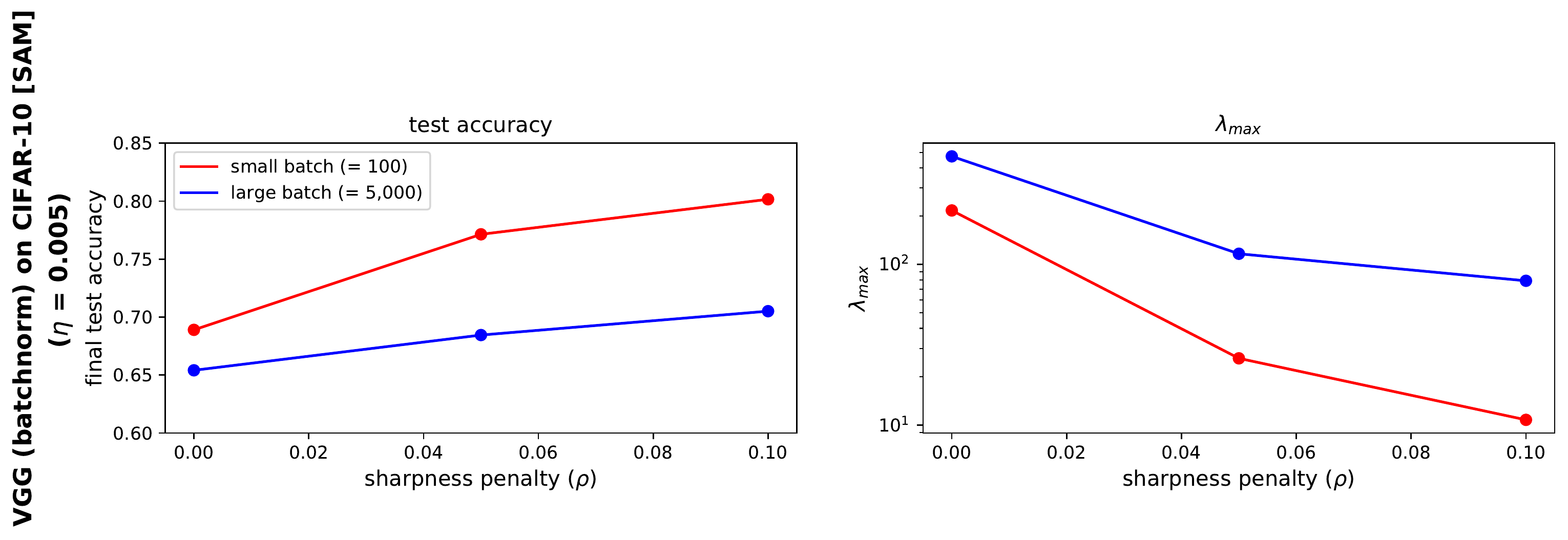}
    \includegraphics[width=\textwidth]{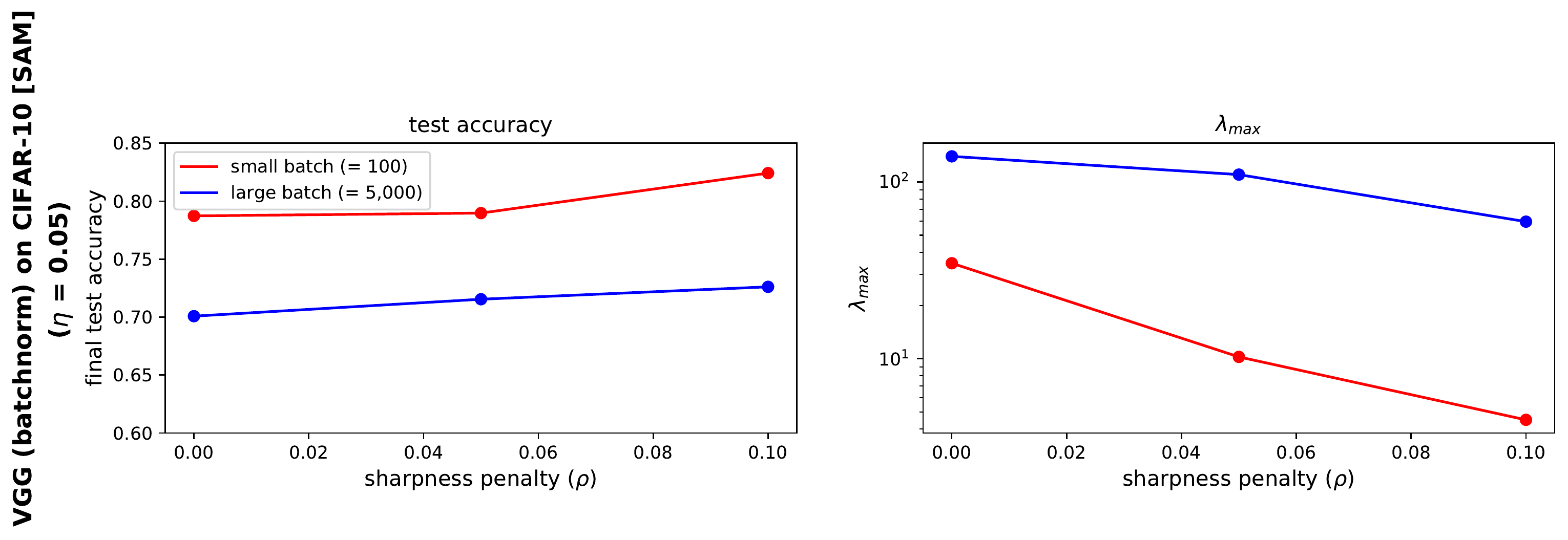}
    \caption{
    \textbf{SAM exhibits generalization benefits for small batch and large batch sizes for BN networks.}
    We train a VGG11 (with BN) using SAM until achieving 99\% train accuracy at a range of values for $\rho$ with learning rates of 0.005 and 0.05. We plot the final test accuracy and \mhe\ (over 1 run with a fixed seed) as a function of $\rho$. Observe that as $\rho$ increases, the final test accuracy increases for both small batch and large batch. In both cases, as $\rho$ increases, \mhe\ decreases.}
    \label{fig:vgg-bn-sam}
    \vspace{-8px}
\end{figure}

We note that large learning rates and SAM can improve final test accuracy in the large-batch regime in the case of BN networks (see Figure \ref{fig:vgg-bn-sam}).

\clearpage
\subsection{Scaling learning rates in the small- and large-batch regimes (continued)}\label{appendix:d}
We run our experiments from Section \ref{sec:sgd} on a modified VGG11 (no BN), where we use GeLU in place of ReLU as the activation function. 
Observe that as the learning rate increases,
the final test accuracy increases for small-batch 
and eventually decreases for large-batch SGD.
In both cases, as the learning rate increases,
\mhe\ decreases.
These findings are consistent with those from Section \ref{sec:sgd}.

\begin{figure}[!ht]
    \centering
    \includegraphics[width=\textwidth]{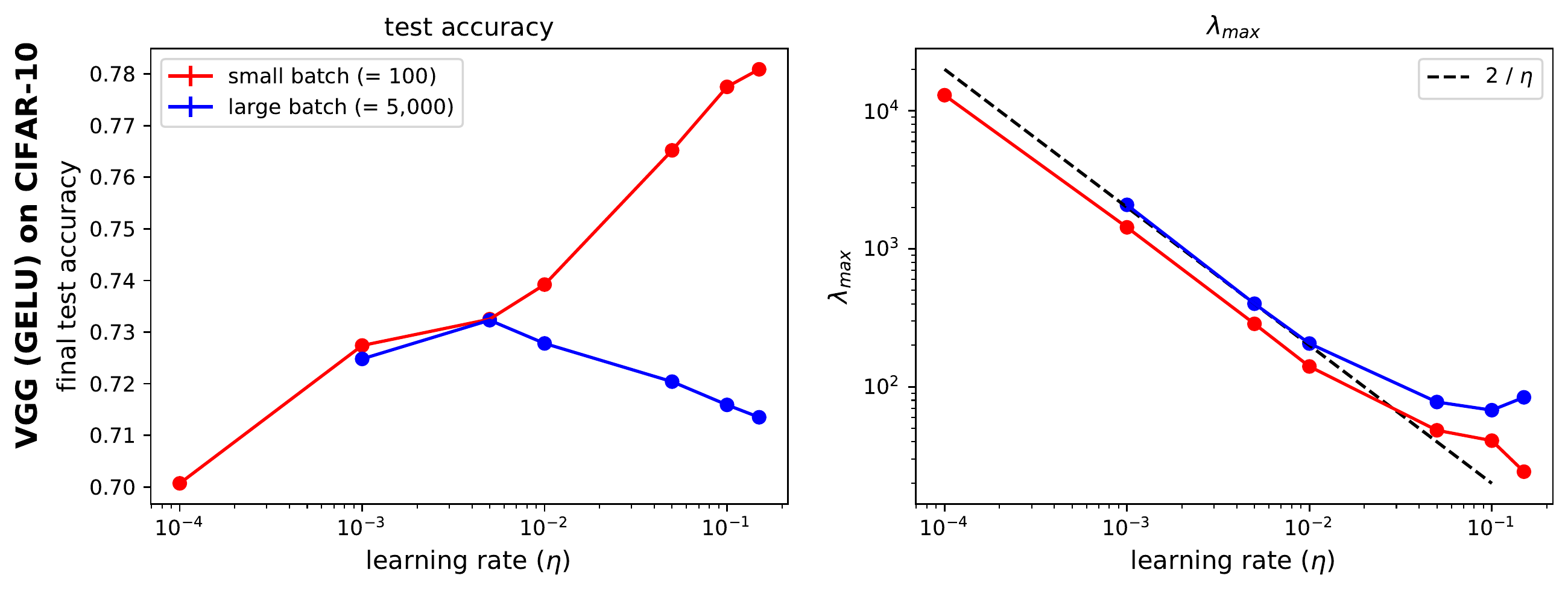}
    \caption{
    We use the same setup as in Figure \ref{fig:vgg-main}, but with a modified VGG11 architecture, where all instances of ReLU as the activation function are replaced with GeLU.
    The results are consistent with Figure \ref{fig:vgg-main}.}
    \label{fig:vgg-gelu}
\end{figure}
\end{document}